\documentclass[10pt,twocolumn,letterpaper]{article}

\usepackage{cvpr}              

\definecolor{cvprblue}{rgb}{0.21,0.49,0.74}
\usepackage[pagebackref,breaklinks,colorlinks,allcolors=cvprblue]{hyperref}
\hypersetup{colorlinks=true,linkcolor=red,urlcolor=magenta}

\def\paperID{10990} 
\def\confName{CVPR}
\def\confYear{2025}
\usepackage{multirow}
\usepackage{tabularx}
\usepackage{pifont}
\title{V2V3D: View-to-View Denoised 3D Reconstruction for Light-Field Microscopy}

\author{Jiayin~Zhao$^{1,2}$\footnotemark[1]~~,~~~Zhenqi~Fu$^{1}$\footnotemark[1]~~,~~~Tao~Yu$^{1}$\footnotemark[2]~~,~~~Hui~Qiao$^{1,2}$\footnotemark[2]\\ 
$^{1}$Tsinghua University, Beijing 100084, China \quad $^{2}$Shanghai AI Laboratory, China\\
{\tt\small zhao-jy23@mails.tsinghua.edu.cn, \{fuzhenqi, ytrock, qiaohui\}@mail.tsinghua.edu.cn}
}

\begin{document}
\maketitle
\renewcommand{\thefootnote}{\fnsymbol{footnote}}
\footnotetext[1]{Co-first author}
\footnotetext[2]{Corresponding author}
\renewcommand{\thefootnote}{\arabic{footnote}}
\begin{abstract}

Light field microscopy (LFM) has gained significant attention due to its ability to capture snapshot-based, large-scale 3D fluorescence images. 
However, existing LFM reconstruction algorithms are highly sensitive to sensor noise or require hard-to-get ground-truth annotated data for training. 
To address these challenges, this paper introduces V2V3D, an unsupervised view2view-based framework that establishes a new paradigm for joint optimization of image denoising and 3D reconstruction in a unified architecture.
We assume that the LF images are derived from a consistent 3D signal, with the noise in each view being independent. This enables V2V3D to incorporate the principle of noise2noise for effective denoising.
To enhance the recovery of high-frequency details, we propose a novel wave-optics-based feature alignment technique, which transforms the point spread function, used for forward propagation in wave optics, into convolution kernels specifically designed for feature alignment. 
Moreover, we introduce an LFM dataset containing LF images and their corresponding 3D intensity volumes.
Extensive experiments demonstrate that our approach achieves high computational efficiency and outperforms the other state-of-the-art methods. These advancements position V2V3D as a promising solution for 3D imaging under challenging conditions.
Our code and dataset will be publicly accessible at \url{https://joey1998hub.github.io/V2V3D/}.

\end{abstract}

\section{Introduction}

\begin{figure}[t]
    \centering
    \setlength{\abovecaptionskip}{0cm} 
    \setlength{\belowcaptionskip}{-0.4cm}
    \includegraphics[width=\linewidth]{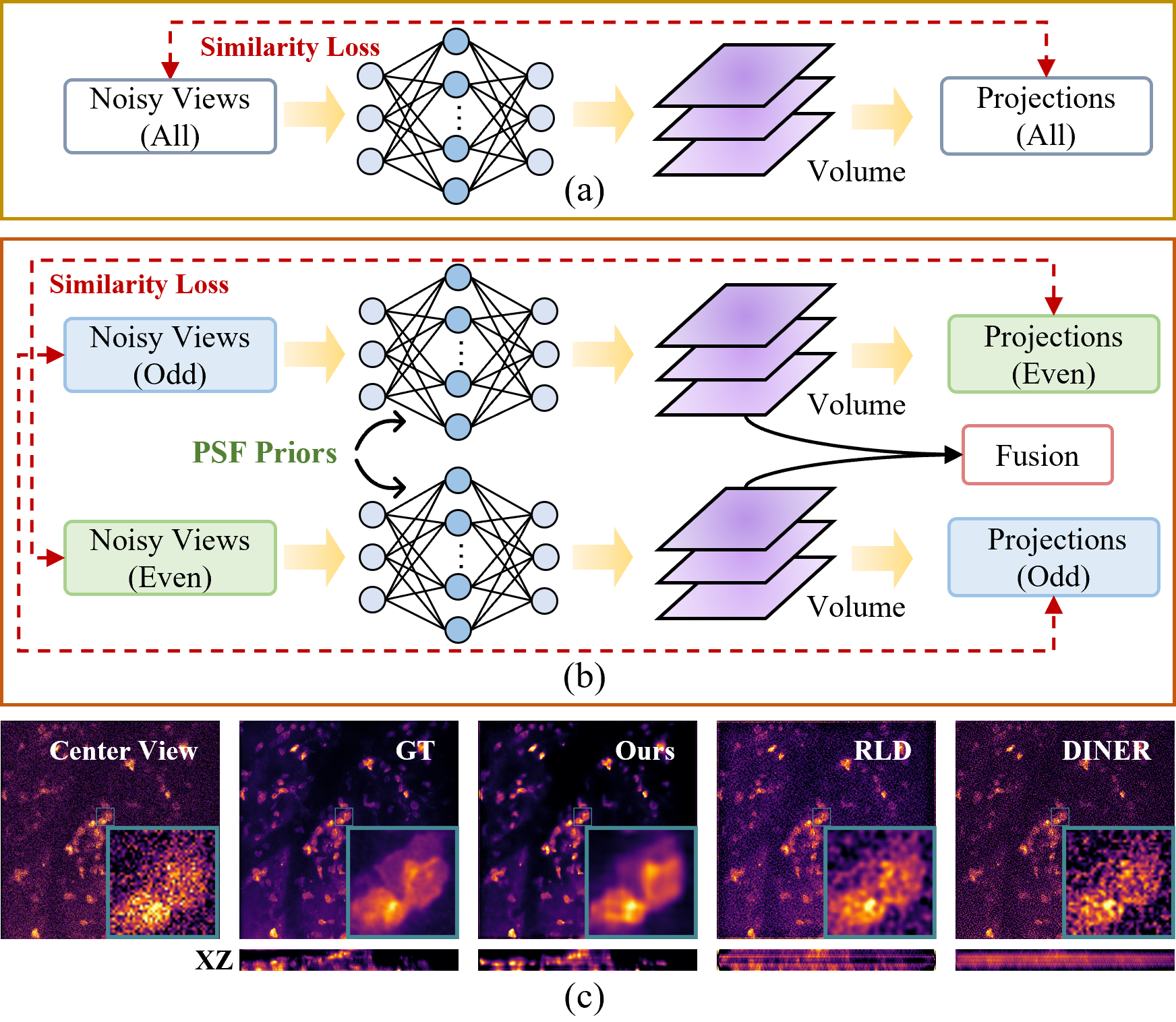}
    \caption{Background and concept of V2V3D. (a) Previous methods, such as VCDNet~\cite{vcdnet} and DINER~\cite{DINER}, directly apply all views for reconstruction and lack physical priors in feature representation. Therefore, when reconstructing real-world noisy scenes, these methods usually generate results with conspicuous artifacts and blurriness.
    (b) The proposed method divides the noisy views into two non-overlapping subsets and employs two networks to generate the corresponding volumes. Additionally, we incorporate PSF priors for feature alignment, thereby enhancing feature aggregation across views. (c) Through the aforementioned custom designs, our method achieves state-of-the-art performance.}
    \label{fig_v2v}
\end{figure}

Light field microscopy (LFM) has emerged as an critical technology in a diverse array of biomedical applications, due to its unparalleled ability to capture high-resolution, three-dimensional microscopic scenes with exceptional precision and efficiency~\cite{pegard2016compressive,wu2021iterative}.
The ability of simultaneously recording both spatial and angular information from the sample allows LFM to generate volumetric data, which is particularly useful in dynamic biological processes where depth and temporal resolution are both crucial~\cite{LFcao,guo2019fourier,rush3d}.

The most classical LFM reconstruction methods can generally be divided into two main categories: Richard-Lucy deconvolution-based (RLD-based) approaches and learning-based solutions. Specifically, RLD-based methods ~\cite{laasmaa2011application,lu2019phase} rely on a computationally expensive and iterative recovery process, which severely limits the overall throughput of LFM reconstruction. This limitation makes them less suitable for long-duration or real-time applications. 
With the rapid progress of deep learning, learning-based algorithms~\cite{DINER,RLN,vcdnet}, predominantly supervised learning methods, have emerged to enhance the speed and quality of LFM reconstruction. However, due to limited generalization capabilities, these algorithms are more suitable for scene-specific reconstruction.

Moreover, fluorescence microscopy of live cells requires gentle conditions, often necessitating low-light conditions that can result in substantial noise~\cite{qu2024self}. 
Existing LFM reconstruction algorithms directly utilize all available views to guide training.
These methods implicitly average the noise across limited views, resulting in noticeable artifacts in the output.
Using the pixel-wise independence of noise~\cite{N2N}, noise-to-noise-based (N2N) methods can effectively reduce noise by learning a mapping between coordinate-matched image pairs.
For light field images (LFIs) with significant noise, previous methods~\cite{2pSAM,wu2021iterative} typically employ a N2N-based temporal denoising method~\cite{li2021unsupervised} to obtain high signal-to-noise ratio (SNR) LFIs before reconstruction.
However, this solution is suboptimal as it may lead to reduced temporal resolution and requires substantial temporal data, making it entirely unsuitable for snapshot applications.
In fact, methods that generate paired noisy images from adjacent frames~\cite{deepsemi,li2021unsupervised} or adjacent pixels~\cite{noise2self,SRDtrans} inevitably result in a substantial reduction in either temporal or spatial resolution~\cite{qu2024self}.

In contrast to existing reconstruction methods that utilize pixel-to-pixel paired images for pre-denoising, we propose a view2view-based simultaneous denoising and 3D reconstruction framework, named V2V3D, which employs view-to-view paired noisy images (pixel-unmatched) as inputs and outputs.
The key to achieving view-to-view denoising lies in establishing a mapping from one image space to another image space, which is physically consistent with the pipeline of unsupervised LFM 3D reconstruction~\cite{DINER}. As illustrated in Figure~\ref{fig_v2v}, the proposed V2V3D splits the views into two subsets, with each subset processed by a separate network to generate the corresponding volumes for fusion. Self-supervised losses are employed between the two branches to facilitate both reconstruction and denoising. 
Furthermore, V2V3D incorporates PSF priors for feature alignment, effectively warping coordinate-unmatched features into coordinate-matched features. This improvement enhances feature aggregation across different views, thereby further boosting the reconstruction performance in detail-rich areas.
Extensive experiments demonstrate that V2V3D outperforms state-of-the-art methods in both noise removal and detail preservation, positioning it as a promising solution for robust snapshot 3D imaging across both microscopic and macro-scale scenarios. The main features of V2V3D are summarized as follows:

\begin{itemize}
    \item A view2view-based simultaneous denoising and 3D reconstruction framework: V2V3D splits all views into two non-overlapping subsets and utilizes two separate networks to reconstruct the corresponding volumes. Using physical priors, it performs forward projection into simulated views, ensuring that the input views and the supervision views are different subsets. The network is optimized by minimizing the differences between the projected views and the real-captured views.
    \item A novel wave-optics-based feature alignment approach: 
    We transform the PSF used in wave optics for forward propagation into convolution kernels for feature alignment, while also eliminating the blurring effects of the PSF. This feature alignment method enables efficient feature aggregation across different views, thereby supporting the recovery of high-frequency details.
    \item A light field dataset for quantitative evaluation: The ground-truth 3D intensity volumes, acquired via fluorescence microscopy, consists of 1618 high-resolution focal stacks.
    Then we utilize the principle of 2pSAM~\cite{2pSAM} to generate the corresponding LF images, as it has been validated to provide high-resolution imaging of deep tissues, particularly in terms of axial resolution. 
\end{itemize}
\section{Related Work}

\begin{figure*}[ht]
    \centering
    \setlength{\abovecaptionskip}{0.2cm} 
    \setlength{\belowcaptionskip}{-0.2cm}
    \includegraphics[width=\linewidth]{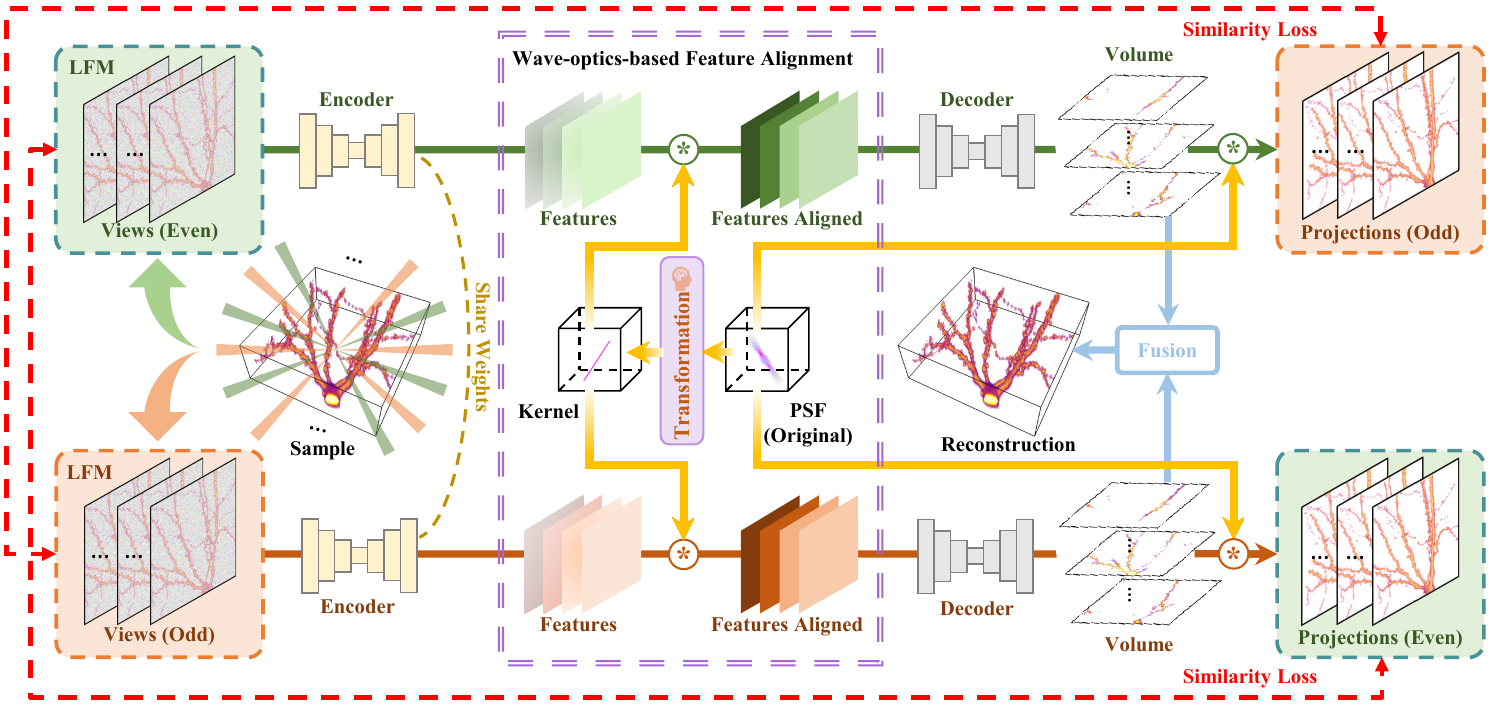}
    \caption{The overall framework of V2V3D, which divides all views into two subsets, with each subset generating a corresponding volume that collaborates to effectively reduce noise. $\circledast$ denotes the 2D convolution operation. Additionally, V2V3D incorporates a novel wave-optics-based feature alignment technique, leveraging PSF priors to enhance the recovery of high-frequency information.}
    \label{fig_pipeline}
    \vspace{-0.5em}
\end{figure*}

\subsection{Traditional LFM 3D Reconstruction}
Light field imaging was initially applied to macro scenarios~\cite{ng2005light}. Due to the ability of LF cameras to simultaneously capture spatial and angular information, they are commonly utilized in tasks such as 3D reconstruction~\cite{codedLF,dynamicLF}, image super-resolution~\cite{cheng2021lightSR,wang2022depth,gao2023spatial}, and depth estimation~\cite{opal,OAVC,jin2020deep}.
Its application in microscopy began in 2006~\cite{levoy2006light}. 
Then, Broxton \etal~ introduced wave optics to model the PSF of LFM ~\cite{broxton2013wave} and applied the RLD method~\cite{fish1995blind} to LFM.
In 2019, Lu \etal~ proposed a  RLD-based method~\cite{lu2019phase} in phase space for LFM, effectively enhancing the reconstruction quality and convergence speed. Furthermore, Wu \etal~ proposed a unique scanning approach in LFM, simultaneously enhancing spatial and angular resolution with reduced phototoxicity~\cite{wu2021iterative}. 
However, conventional fluorescence microscopy struggles to achieve near-diffraction-limited resolution in deep tissue due to refractive index inhomogeneity and scattering~\cite{zipfel2003nonlinear,kong2023neuron}. Two-photon microscopy (TPM) overcomes these issues through its longer wavelength and localized nonlinear excitation~\cite{TPM}. A recent innovation, 2pSAM~\cite{2pSAM}, combines TPM with angular-scanning LF measurement to achieve near-diffraction-limited imaging with reduced photodamage.

However, these advanced LF imaging frameworks rely on RLD-based reconstruction algorithms, which limit the practicality of LFM.
These algorithms iteratively correct the reconstructed volume based on Poisson assumption, essentially averaging the noise from all views in the output. This results in significant artifacts and image smoothing, while the iterative approaches are also computationally inefficient.
Although more sophisticated RLD-based method has emerged for structured illumination microscopy~\cite{zhao2022sparse}, its reliance on sparse priors and the use of strong regularization restrict their applicability to all microscopic contexts.

\subsection{Deep-learning-based LFM 3D Reconstruction}
Although recent deep-learning-based LFM reconstruction methods~\cite{RLN,vcdnet} significantly improve the reconstruction efficiency compared with the RLD-based methods, their reconstruction quality is still far from practical due to poor generalization capacity and the lack of real-captured high-resolution training data. 
Additionally, convolutional networks such as VCDNet~\cite{vcdnet} concatenate all views as separate channels, where significant intensity variation occurs at the same location.
Due to the limited receptive field of convolutional networks, this results in ineffective information aggregation and complicates the recovery of high-frequency details.
Since 2020, implicit neural representation (INR) becomes a hot tool in the computer vision and graphics community for its superior performance on tasks like novel view synthesis~\cite{ying2023parf,mildenhall2021nerf,barron2021mip}, 3D reconstruction~\cite{microdiffusion,liu2024finer,pamir} and physical simulation~\cite{qiao2022neuphysics,wanghx}. 
In recent developments, INR-based methods have emerged in the field of LFM. For example, DeCAF~\cite{decaf} has demonstrated the ability to eliminate the missing cone problem, it is too slow for use in long-term observation (e.g., DeCAF needs 20 hours to reconstruct single volume).
DINER~\cite{DINER} significantly enhances reconstruction accuracy compared to DeCAF. However, its efficiency is still relatively lower when compared to convolutional networks, and it performs poorly in handling noise.

\subsection{Denoising for Microscopy}
Fluorescence microscopy of live cells necessitates gentle imaging conditions and sufficient spatiotemporal resolution, often resulting in a limited photon budget~\cite{traditionaldenoise2010,traditionaldenoise2018,superviseddenoise2018, qu2024self}. To compensate for this constraint, improving the SNR is crucial for accurate LFM reconstruction. 
However, obtaining a sufficient collection of clean images for supervised learning poses significant challenges, particularly in live-cell applications.
To remove noise without clean images, N2N-based methods~\cite{N2N,noise2self,noise2void,huang2021neighbor2neighbor} learn mappings between pairs of independently degraded versions of the same image, achieving performance comparable to supervised methods. 
N2N-based methods have appeared in the field of TPM~\cite{li2021unsupervised,SRDtrans}.
For example, DeepCAD~\cite{deepcad,deepcadrt} employs a self-supervised data generation process that assumes adjacent frames in a continuous imaging video share the same underlying content. However, in light field measurements, the considerable differences between adjacent views pose a challenge for N2N-based methods, which rely on the coordinate-matched image pairs. 
The variation in intensity at the same location can lead to noticeable artifacts.
\section{Method}
In this section, we first provide a brief introduction of LFM 3D imaging. Next, we detail the view-to-view framework and the feature alignment mechanism, emphasizing their crucial role in both denoising and reconstruction. Finally, we present the network architecture and loss functions.

\subsection{Preliminaries}
LFM employs 2D angular scanning techniques, such as LED multi-angle illumination and microlens arrays, to achieve high-speed 3D imaging with subcellular resolution.
To simulate real-world LFM imaging, we developed a mathematical model that captures the entire process of light field imaging.
First, we derive the point spread function (PSF) representation by modeling the light propagation process within a wave optics framework. This model encompasses the entire journey from the laser output to the objective plane.
We define the direction of light propagation as the Z-axis and sample $z$ points, with the intensity of each point represented as $I_{x,y,z}$. As illustrated in Figure~\ref{fig_pipeline}, there are $U$ beams illuminating the sample from different angles, each modeled as a PSF, denoted as $PSF_{u,x,y,z}$. 
Then the captured light field image (LFI) is represented as:
\begin{equation}
LFI_{u,x,y} = \sum\nolimits_z (I_{x,y,z}*PSF_{u,x,y,z}), 
\label{eq:fowardprojection}
\end{equation}
where $*$ denotes the 2D convolution operation.

\subsection{View-to-View-based LFM 3D Reconstruction}
Fluorescence microscopy of live cells requires gentle conditions, often necessitating low-light environments that can result in substantial sensor noise. Due to the absence of noise modeling, current LFM reconstruction algorithms typically average noise across views in their results. However, this approach is ineffective at managing severe noise and may introduce significant artifacts.
The N2N-based denoising methods leverage the inherent property of neural networks to avoid generating random noise. This characteristic facilitates effective noise reduction by mapping between pairs of coordinate-matched, noise-independent degraded versions of images, thereby preserving the underlying consistent signals.
However, as illustrated in Eq.~\ref{eq:fowardprojection}, the projection matrices (i.e., PSFs) that map the 3D information to 2D space differ for each view. 
As a result, the 2D coordinates of photons emitted from the same point in a 3D sample differ across different views, preventing the direct application of N2N-based methods for noise removal.

In this study, we propose a view2view-based framework that incorporates the principle of N2N~\cite{N2N} for denoising, enabling the reconstruction of high-quality 3D signals without ground truth data. 
We assume that the LFIs are fundamentally derived from a consistent 3D signal, with the noise in each view being independent.
Our approach is to reconstruct a 3D signal using information from several views and then generate the remaining views using Eq.~\ref{eq:fowardprojection}). 
This process generates pairs of coordinate-matched and noise-independent LFIs, thereby satisfying the N2N assumptions for effective denoising. 
Specifically, we divide all $U$ views equally into two non-overlapping subsets, $U_1$ and $U_2$.
This partitioning strategy offers two key advantages: 1)
All views are engaged per iteration, with one subset as input and the other for supervision; 2) The disjoint input/output pairing prevents trivial identity mappings, compelling the network to generate noise-free views by optimizing towards the statistical expectation of the target distribution.

As illustrated in Figure~\ref{fig_pipeline}, the simulated view $u_2$ in subset $U_2$ can be generated using the subset $U_1$, as expressed by
\begin{equation}
\hat{LFI}_{u_2,x,y} = \sum\nolimits_z(f(LFI_{U_1,x,y}) * PSF_{u_2,x,y,z}),
\end{equation}
where $f(\cdot)$ indicates a U-Net for reconstructing $\hat{I}_{x,y,z}$. Similarly, the simulated view $u_1$ in subset $U_1$ can be generated using the subset $U_2$, as expressed by
\begin{equation}
\hat{LFI}_{u_1,x,y} = \sum\nolimits_z(f(LFI_{U_2,x,y}) * PSF_{u_1,x,y,z}).
\end{equation}
This network is optimized by minimizing the difference between the real-captured and simulated LFIs, yielding a model capable of high-quality reconstruction and denoising. Moreover, we implement two branches for reconstruction and merge the 3D signals produced by both branches to obtain the final reconstruction result.

\begin{figure}[t]
    \centering
    \setlength{\abovecaptionskip}{0.2cm} 
    \setlength{\belowcaptionskip}{-0.4cm}
    \includegraphics[width=\linewidth]{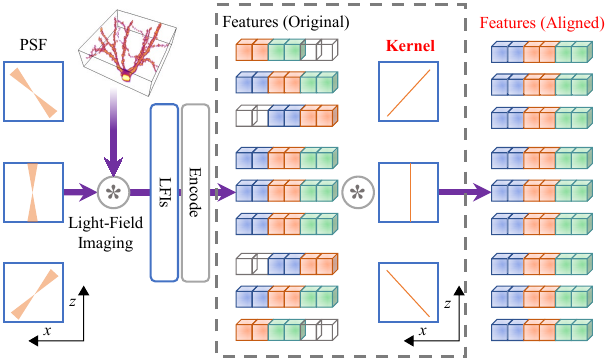}
    \caption{The diagram of the proposed wave-optics-based feature alignment module. The features extracted from different views are misaligned in the spatial dimension. To address this, we use kernels generated from the PSFs to align these features, thereby facilitating subsequent feature aggregation.}
    \label{fig_feature}
\end{figure}

\begin{table*}[!t]
    \centering
    \setlength{\abovecaptionskip}{0.1cm} 
    \renewcommand{\arraystretch}{1.2}
    \setlength{\tabcolsep}{6.6pt}
    \small
    \caption{Quantitative comparison with state-of-the-art methods on the synthetic dataset. The best results are highlighted in bold.}
    \begin{tabular}{c|cc|cc|cc|cc|cc|cc} %
        \hline
        \multirow{2}*{Scene} & \multicolumn{2}{c|}{RLD} & \multicolumn{2}{c|}{VCDNet} & \multicolumn{2}{c|}{DINER} & \multicolumn{2}{c|}{DeepCAD+RLD} & \multicolumn{2}{c|}{DeepCAD+DINER}& \multicolumn{2}{c}{Ours} \\
        \cline{2-13}
        & PSNR & SSIM & PSNR & SSIM   & PSNR & SSIM  & PSNR & SSIM & PSNR & SSIM  & PSNR & SSIM  \\
        \hline
        B cells & 27.13 & 0.507  & \textbf{45.15} & 0.972 & 28.01 &0.481 & 29.05& 0.896& 32.58&  0.859& 38.73 & \textbf{0.981} \\
        Dendrites 1 & 27.30 & 0.516  & 30.41 & 0.709 & 26.85 & 0.386 &30.16 & 0.734& 30.47& \textbf{0.806}& \textbf{30.92} & 0.768\\
        Dendrites 2  & 34.76 & 0.819  & 36.25 & 0.851 &33.14  & 0.740  & 34.88& 0.866& 34.82& 0.857& \textbf{36.65} & \textbf{0.897} \\
        Neutrophils & 34.28 & 0.783  & 30.85 & 0.555  & 26.59 & 0.345  & 36.56& 0.787& 36.38& \textbf{0.928}& \textbf{39.94} & 0.851 \\
        Microglia 1 & 30.43 & 0.800   & \textbf{40.16} & 0.960  & 26.52 & 0.560 & 32.15& 0.895& 35.53& 0.958&38.35 & \textbf{0.966}\\
        Microglia 2 & 30.41 & 0.789  & \textbf{40.84} & 0.954 &28.42  & 0.644&32.44 &0.902 & 35.36& 0.940& 39.19 & \textbf{0.964} \\
        Neurons 1 & 31.52 & 0.697  & \textbf{37.44} &0.747 & 31.12 & 0.625  & 33.19& 0.821&33.17 & 0.825& 36.39 & \textbf{0.783} \\
        Neurons 2 & 29.60 & 0.586& 32.74 & 0.631  & 28.69 & 0.504 & 31.57& 0.761& 30.75& 0.707& \textbf{32.92} & \textbf{0.806} \\
        Neurons 3 & 37.39 & 0.839  & 42.85 & 0.870 & 36.70 & 0.795 & 40.70&0.920 &39.72 & 0.921& \textbf{43.30} & \textbf{0.932} \\
        Vessels 1 & 34.86 & 0.758 & 42.02 & 0.817 & 30.40 & 0.482  & 42.18&0.926 &42.22 &0.952 & \textbf{44.29} & \textbf{0.952}\\
        Vessels 2& 36.46 & 0.838 & 36.06 & 0.645 & 29.23 & 0.459 & 46.63& 0.964& 40.96&0.977 & \textbf{48.91} & \textbf{0.976} \\
        \hline
        Average & 32.19& 0.721 & 37.71 & 0.792  & 29.60 & 0.547 & 35.41& 0.861& 35.63&0.885 & \textbf{39.05} & \textbf{0.898} \\
        \hline
    \end{tabular}
    \label{tab:compare}
    \vspace{-1em}
    
\end{table*}

\subsection{Wave-optics-based Feature Alignment}
The primary reason for the insufficient reconstruction quality of convolutional networks such as VCDNet is that they concatenate all views as separate channels and directly input them into the network.
Due to the limited receptive field of convolutional networks, significant intensity variations across different channels at the same location can hinder effective information aggregation and impede the enhancement of reconstruction performance in detail-rich areas.

In RLD-based methods~\cite{laasmaa2011application,lu2019phase}, the information from the error map calculated between the simulated and real-captured LFIs can be progressively updated into the reconstructed 3D signal through back-projection, as expressed by
\begin{equation}
{\Delta}I_{update} = Error(LFI,\hat{LFI}) * PSF^{-1},
\label{eq:backprojection}
\end{equation}
where $PSF^{-1}$ can be obtained by flipping the $PSF$ in two-dimensional space, as expressed by
\begin{equation}
PSF^{-1}_{u,x,y,z} = PSF_{u,-x,-y,z}.
\end{equation}
Inspired by the back-projection technique used in RLD, we propose a wave-optics-based feature alignment method that enhances effective feature aggregation across different views.
After extracting features for each view, we apply back-projection to warp all feature maps from their respective 2D spaces into a unified 3D space, facilitating improved feature aggregation.
One might consider directly using $PSF^{-1}$ for the back-projection of the feature maps. However, since the PSF acts as a low-pass filter in the frequency domain, this approach inevitably leads to feature blurring.
To mitigate blurring effects, the PSF is converted into a convolution kernel with diameter 1 and weight 1, as illustrated in Figure~\ref{fig_feature}. Specifically, we compute centroid coordinates of each PSF slice, set their values to 1, and set non-centroid positions to 0.
Then the entire feature alignment process can be expressed as
\begin{equation}
Feature_{align} = Feature * Kernel_{PSF^{-1}}.
\label{eq:featalign}
\end{equation}

\subsection{Network Architecture}
As shown in Figure~\ref{fig_pipeline}, the input of our method is the real-captured LFIs, and the output is a high-resolution 3D intensity volume. The overall V2V3D reconstruction framework comprises two branches. Each branch includes: I) An encoder with a pyramid structure for feature extraction, with weights shared between two branches; II) A wave-optics-based feature alignment module that utilizes back-projection to warp all feature maps from different 2D spaces into a unified 3D space, facilitating feature aggregation; III) A U-Net-based decoder for generating a 3D volume from the aligned feature maps; IV) A forward projection module based on the physical modeling of the LFM system to produce simulated LFIs.
The final reconstruction result is obtained by averaging the two volumes generated by the branches. Further details of the network architecture can be found in the supplementary materials.

\subsection{Loss Functions}
Although the MSE loss performs well in most scenarios, its effectiveness diminishes significantly in LFM reconstruction due to optical defocus, resulting in oversmoothing of high-frequency details.
Therefore, we adopt the FFT Loss~\cite{zhao2024pnr} to better recover high-frequency details.
By utilizing the fast Fourier transform to map images from the spatial to the frequency domain, the FFT loss effectively balances the optimization of information across various frequencies.
We define $LFI_{i}$ as the value of real-captured projection pixel $i$, with the corresponding estimated pixel value denoted as $\hat{LFI}_i$. The MSE loss is defined as 
\begin{equation}
L_{MSE} = \frac{\sum_{i}(LFI_{i} - \hat{LFI}_i )^2}{N},
\end{equation}
while the FFT loss can be expressed as
\begin{equation}
L_{FFT} = \frac{\left\| {FFT(LFI) - FFT(\hat{LFI}}) \right\|_2^2}{N},
\end{equation}
where $FFT(\cdot)$ indicates the fast Fourier transform and $N$ is the total number of the pixels. 

We also designed a regularization loss to mitigate artifacts caused by signal crosstalk along the Z-axis. In the reconstruction results of LFIs with high brightness, a slice may exhibit excessively high intensity, while other slices could appear too dark, potentially falling below the sensor's background noise level. This discrepancy can lead to the emergence of significant artifacts.
Therefore, we apply the de-crosstalk loss to penalize values that fall below the background noise level of the sensor.
Specifically, the de-crosstalk loss is defined as
\begin{equation}
L_{DC} = \sum\nolimits_{x,y,z}ReLU(BG-\hat{I}_{x,y,z}),
\end{equation}
where $BG$ is the intensity of background noise, which can be estimated based on histogram analysis of all LFIs. 

The final loss function is composed of three components: the MSE loss $L_{MSE}$, the FFT loss $L_{FFT}$, and the de-crosstalk loss $L_{DC}$. 
Thus, the overall training loss $L_{all}$ is expressed as
\begin{equation}
L_{all} = L_{MSE} + \alpha L_{FFT} + \beta L_{DC},
\end{equation}
where $\alpha$ and $\beta$ are weights. 
Empirically, we set $\alpha=0.1$ and $\beta=1$.
\section{Experiments}
\begin{table}[t]
    \centering
    \setlength{\abovecaptionskip}{0.1cm} 
    \setlength{\tabcolsep}{7.5pt}
    \small
    \caption{Efficiency comparison of V2V3D with other methods.}
	\begin{tabular}{ccccc}
		\hline
		Method & PSNR & SSIM & Runtime (s) & Params (M)\\
		\midrule         
		VCDNet & 37.71 &  0.792  & 0.047 & 87.98\\
		RLD &  32.19& 0.721 & 7.34 & - \\
		DINER  & 29.60 & 0.547 & 61.4 & 62.92\\
    	V2V3D   & 39.05 & 0.898 & 0.413 & 210.77\\   
		\bottomrule
    \end{tabular}
    \label{tab:Efficiency}
    \vspace{-1.8em} 
\end{table}
In this section, we first introduce the experimental setup and datasets for evaluation. 
Then, we present both quantitative and qualitative comparisons with other SOTA methods. 
Finally, we conduct ablation studies to examine the various components of V2V3D.

\subsection{Experimental Setup and Datasets}
We confirmed the superior performance of V2V3D using both synthetic and real-world data. 
We utilized the principle of 2pSAM to obtain LFIs, as illustrated in Figure~\ref{fig_pipeline}. 
This system features a distinctive ``needle" beam for advanced light field imaging, facilitating both 2D spatial and angular scanning. This allows for high-speed, large-field 3D imaging at subcellular resolution. By rotating the mechanism, we can capture 13 LFIs of the 3D sample. 
For the synthetic dataset, we observed six types of biological scenarios using fluorescence microscopy, including B cells, dendrites, microglia, neurons, neutrophils, and blood vessels.
Applying cropping and resizing operations, we obtained 1618 high-SNR 3D intensity volumes, each with a resolution of $512 \times 512 \times 39$. Then we used the generated PSFs of 2pSAM\footnote{\url{https://github.com/BBNCELi/2pSAM_recon}} and intensity volumes to perform physics-based forward projection, resulting in 1618 simulated light field images, each with a resolution of $512 \times 512 \times 13$. 
Finally, we introduced substantial Gaussian noise into the LFIs to simulate the actual imaging process.
We selected 11 typical cases from the dataset to constitute the test set.
For the real-world dataset, we used the 2pSAM system to obtain LFIs. 
Note that we selected thick samples and shortened the exposure time to obtain low-SNR LFIs.
We observed one static sample (brain-slice of mouse) and one live sample (neutrophils). Both of them consist of 100 frames, with each frame having a resolution of $512 \times 512 \times 13$. For the hardware configuration, we utilized an NVIDIA A100 GPU to handle the large-scale data and complex calculations involved in our study. For each method, we used PSNR and SSIM~\cite{wang2004image} to evaluate the accuracy of the reconstruction. 
Following the approach in \cite{wu2021iterative}, we subtract the background noise from both the reconstruction results and the ground truth before calculating the metrics.

\begin{figure*}[t]
	\centering
        \setlength{\abovecaptionskip}{0.2cm} 
        \setlength{\belowcaptionskip}{0cm} 
	\includegraphics[width=1.0\textwidth]{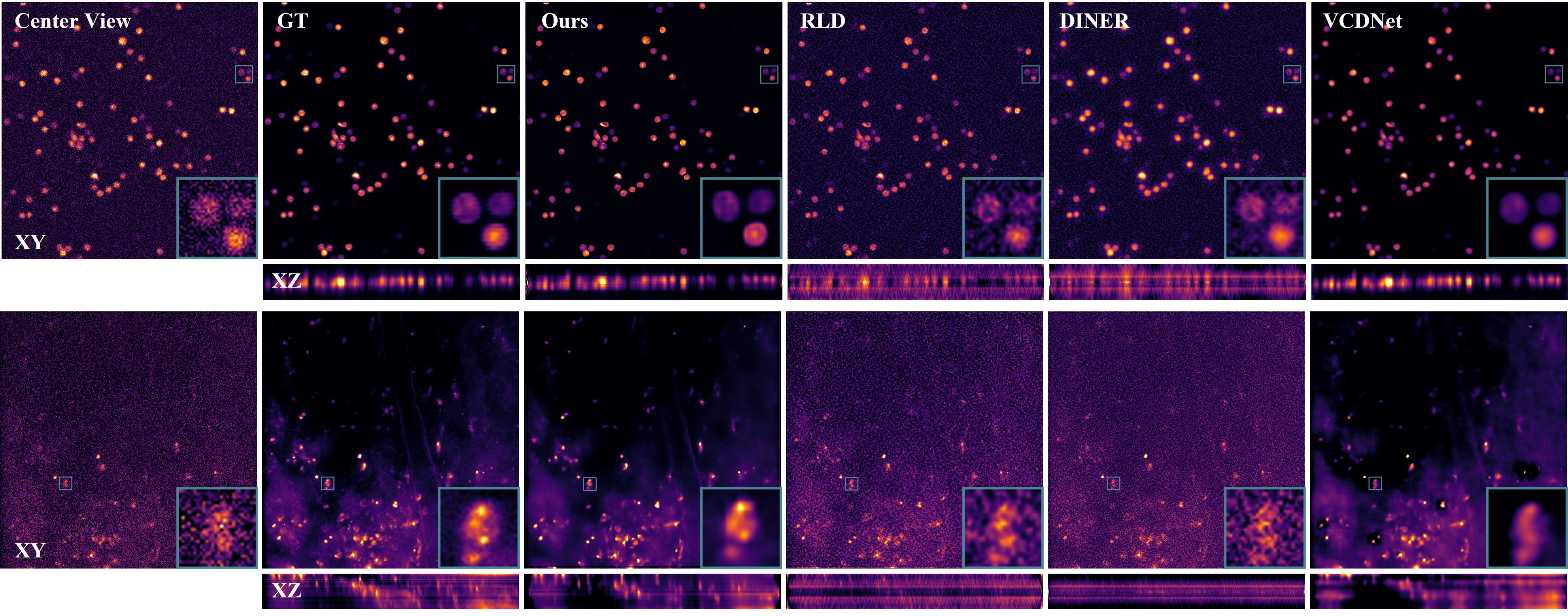} 
    \vspace{-1.8em}
	\caption{Qualitative comparisons on the synthetic dataset. Two biological samples arranged from top to bottom are B cells and vessels. Our solution delivers significantly higher quality, with less noise and sharper details. }
	\label{fig_syn_compare}
    \vspace{-0.8em}
\end{figure*}

\begin{figure*}[t]
	\centering
        \setlength{\abovecaptionskip}{0.2cm} 
        \setlength{\belowcaptionskip}{-0.2cm} 
	\includegraphics[width=1\textwidth]{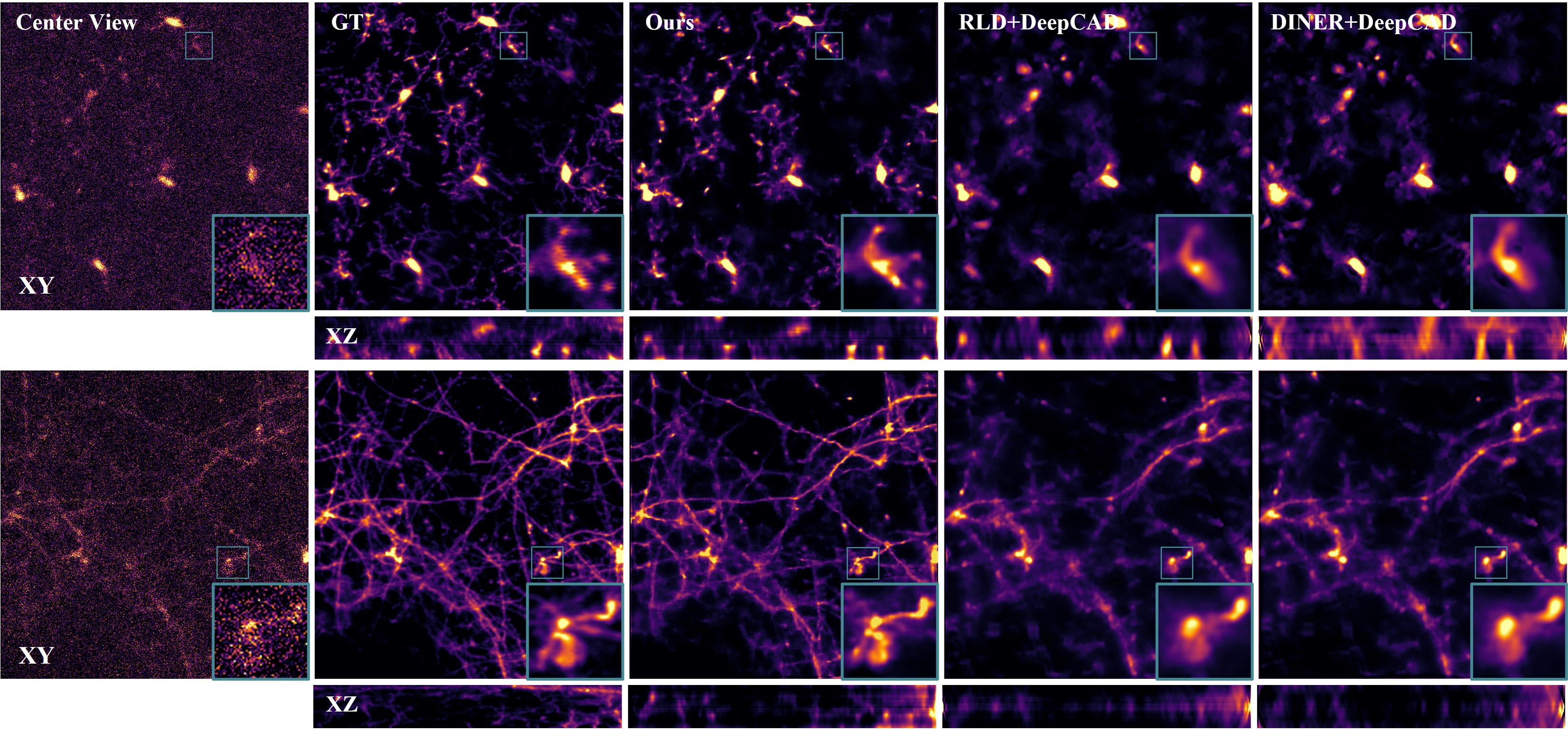} 
    \vspace{-1.5em}
	\caption{Qualitative comparisons on the synthetic dataset. Two biological samples arranged from top to bottom are microglia and dendrites. Our solution delivers significantly higher quality, with less noise and sharper details.}
	\label{fig_syn_deepcad_compare}
    \vspace{-0.6em}
\end{figure*}

\begin{figure*}[t]
	\centering
        \setlength{\abovecaptionskip}{-0.05cm} 
        \setlength{\belowcaptionskip}{-0.15cm} 
	\includegraphics[width=1\linewidth]{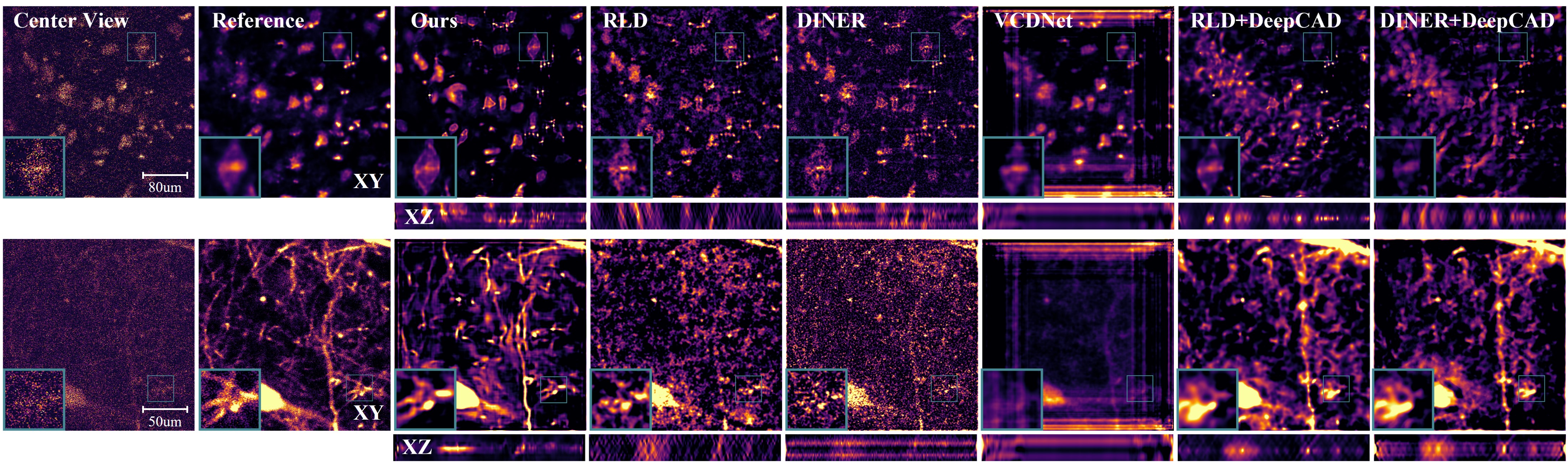} 
	\caption{Qualitative comparisons on the real-world dataset. Two biological samples arranged from top to bottom are neutrophils and dendrites. For live sample (Neutrophils), we employ DeepCAD for denoising to acquire the reference center view. For static sample (Dendrites), we obtained a high SNR reference center view through time averaging.}
	\label{fig_real_compare}
    \vspace{-0.8em}
\end{figure*}

\begin{figure}[t]
    \centering
    \includegraphics[width=1\linewidth]{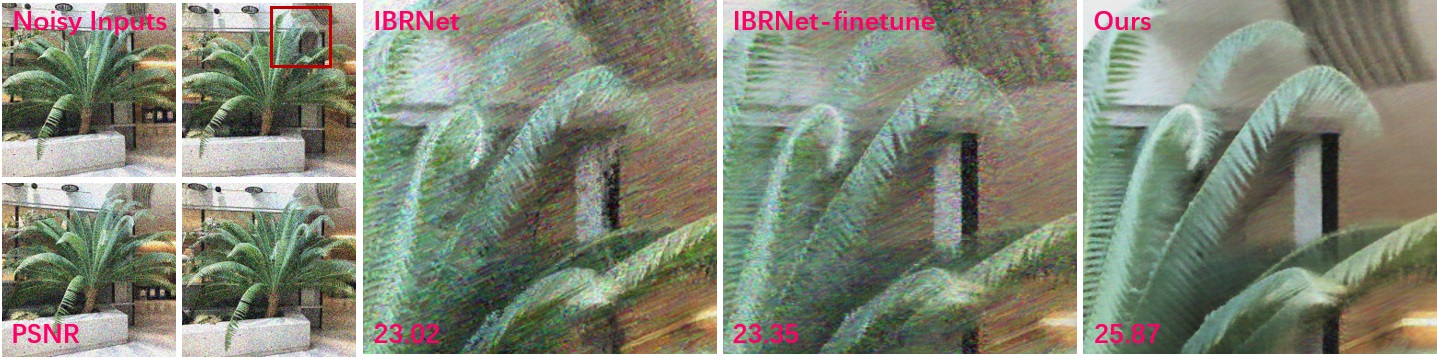} 
    \vspace{-2em}
    \caption{Validation of denoising ability on macro-scale scenario.}
    \vspace{-1.4em}
    \label{fig_IBR}
\end{figure}

\subsection{Comparison on the Synthetic Dataset}
We quantitatively compared our method with other SOTA ones on synthetic data, including an optimization-based method (RLD)~\cite{lu2019phase}, a supervised-learning-based method (VCDNet)~\cite{vcdnet} and a NeRF-based method (DINER)~\cite{DINER}. 
We retrained VCDNet on our dataset, using noisy LFIs as input and the noise-free 3D signal as supervision.
Moreover, to validate the superiority of our framework in simultaneous reconstructing and denoising, we compared V2V3D with the unsupervised methods (RLD and DINER) that use pre-denoised LFIs as input. Note that we retrained DeepCAD~\cite{deepcad} on our dataset for LFI denoising.

Table~\ref{tab:compare} reports the average metrics of all methods applied to noisy and pre-denoised synthetic data. Figure~\ref{fig_syn_compare} and Figure~\ref{fig_syn_deepcad_compare} show the center views as well as the XY and XZ projections of the reconstructed and ground-truth 3D volumes.
Table~\ref{tab:Efficiency} offers a comprehensive quantitative comparison of all methods, including performance metrics and elapsed times. 
We can obtain the following conclusions:
I) On the synthetic dataset, V2V3D demonstrates superior performance, particularly when compared to unsupervised methods. By leveraging the view-to-view reconstruction framework, our method is capable of recovering high-resolution, high-SNR 3D signals from LFIs with severe noise. Additionally, while using denoised LFIs as input can enhance the performance of RLD and DINER, it may also introduce more artifacts into the reconstructed 3D volumes;
II) Due to the use of noise-free 3D signals as supervision, VCDNet can mitigate the impact of noise on reconstruction. However, by leveraging a physics-informed feature alignment module, V2V3D is able to reconstruct more high-frequency details than VCDNet;
III) Benefiting from the convolutional framework, V2V3D shows superior reconstruction efficiency over optimization-based methods. 

\subsection{Comparison on Real-world Dataset} 
We conducted qualitative comparisons using two types of real data: one static sample (brain slice) and one live sample (neutrophils). 
For the static sample, we can obtain a high-SNR reference center view through time averaging (100 frames), while for the live sample, we employed DeepCAD for denoising to acquire a relatively high-SNR reference center view.
Figure~\ref{fig_real_compare} shows the center views, as well as the XY and XZ projections of the reconstructed 3D volumes.
Severely affected by noise, both RLD and DINER exhibited significant deficiencies, as these methods essentially treat noise as valid signals in reconstruction. Consequently, the mean projection of the reconstructed volume closely resembles the center view.
Since VCDNet was trained on a noisy dataset, it is capable of denoising to some extent. However, due to its poor generalization ability, the reconstruction results exhibit noticeable artifacts.
Using pre-denoised LFIs from DeepCAD for reconstruction does improve the SNR of RLD and DINER. However, due to significant differences between adjacent views at the same position, directly applying the N2N-based denoising method can result in signal crosstalk and blurring in LFIs, leading to the emergence of noticeable artifacts in final reconstruction results.
Our method consistently outperformed other SOTA methods through several pivotal factors: the view2view framework effectively separates valid signals from severe noise, while the physics-informed feature alignment and FFT Loss boost the network's capacity to recover high-frequency details. 
These innovative techniques highlights V2V3D's potential applicability in real-world complex LFM imaging.

\subsection{Validation on Macro-scale Scenarios.} 
To validate the denoising ability of the view2view training strategy on macro-scale scenarios, we adapted V2V3D by building upon IBRNet~\cite{wang2021ibrnet}. 
Specifically, we sub-sampled the views of the \textit{Fern} scene in the LLFF~\cite{mildenhall2019local} by a factor of 4 and added severe Gaussian noise ($\mu=0$, $\sigma=50$). Then, we replaced the parts that generate views from one subset to another in both branches of V2V3D with IBRNet, and randomly sampled two non-overlapping subsets of views for training in each iteration.
As shown in Figure~\ref{fig_IBR}, our method demonstrates robust performance on noisy macro images, outperforming both the pre-trained IBRNet and the one fine-tuned specifically for the \textit{Fern} scene.

\subsection{Ablation Study}    
Table~\ref{tab:ablation} presents the contributions of every crucial component within V2V3D.
Further visualizations of the ablation study can be found in the supplementary materials.
Specifically, removing the view2view framework would cause the network to lose its denoising capability, resulting in a substantial drop in performance.
Additionally, eliminating the feature alignment module results in a considerable loss of detail and overall image sharpness, further compromising the effectiveness of the reconstruction process. 
Furthermore, when the de-crosstalk loss is excluded, the artifacts caused by signal crosstalk in the reconstruction results increase significantly, degrading the quality of the output.
Collectively, these findings underscore the critical importance of each component in enhancing the reconstruction quality, highlighting that each element plays a vital role in achieving optimal performance.
We also explored other fusion strategies, e.g., max-pooling and learnable aggregation. Yet, as shown in the supplementary materials, these approaches brought no significant improvement.
\begin{table}[!t]
\centering
\setlength{\abovecaptionskip}{0.1cm} 
\setlength{\belowcaptionskip}{0cm}
 \small
 \setlength{\tabcolsep}{4pt}
\caption{Ablation study on the V2V framework, feature alignment strategy, and the impact of FFT and de-crosstalk losses.}
	\begin{tabular}{cccccc}
		\hline
		Metric & Ours & w/o V2V & w/o Align & w/o $L_{FFT}$ & w/o $L_{DC}$ \\
		\midrule         
		PSNR & 39.05 &30.81  & 38.25 & 37.23 & 36.09  \\
		SSIM & 0.898 & 0.731 & 0.885 & 0.874 & 0.867  \\
		\bottomrule
	\end{tabular}
	\label{tab:ablation}
    \vspace{-1.6em} 
\end{table}

\section{Conclusions}
\noindent \textbf{Limitations and Future Works:} 
Although V2V3D has achieved SOTA performance and shows significant promise in life science, there are still two directions for improvement: I) Develop more advanced fusion strategies rather than using straightforward averaging to further improve performance; II) Incorporate the optimization of the PSF during training to reduce the dependence of unsupervised methods on accurate imaging system models.

\noindent \textbf{Conclusion:} 
This study presents V2V3D, an view2view-based simultaneous denoising and 3D reconstruction framework for LFM. V2V3D divides all views into two non-overlapping subsets, each subset generating a corresponding volume and collaborating to remove noise. We also introduce a novel wave-optics-based feature alignment technique to improve reconstruction accuracy in detail-rich areas. 
Moreover, we introduce an LFM dataset to enable both quantitative and qualitative comparisons.
We believe that V2V3D serves as a seminal exploration in simultaneous denoising and reconstruction, capable of stimulating more research within this burgeoning field.
\section{Acknowledgement}
This work was supported by National Key R\&D Program of China (2023YFB3209700), National Natural Science Foundation of China (62322110, 62422110), and Postdoctoral Fellowship Program of CPSF (GZC20240836).

{\small
\bibliographystyle{ieeenat_fullname}
\bibliography{main}}


\end{document}